\documentclass{article}
\usepackage[accepted]{icml2019}

\RequirePackage{times}

\usepackage{amssymb,amsmath,amscd,amsfonts,amstext,amsthm,bbm, enumerate, dsfont, mathtools}
\usepackage{array}
\usepackage{multicol}
\usepackage{algorithm}
\usepackage{graphicx}
\usepackage{hyperref}
\usepackage{import}
\usepackage{caption,subcaption}
\usepackage{footnote}
\usepackage{diagbox}
\usepackage{placeins}
\usepackage{booktabs}
\usepackage{wrapfig}

\graphicspath{ {figures/} }

\newtheorem{theorem}{Theorem}

\newcommand{\sumin}{\sum_{i=1}^n}

\newcommand{\RR}{\mathbb{R}}

\newcommand{\cN}{{\cal N}}
\newcommand{\Var}{\mathrm{Var}}
\newcommand{\cov}{\mathrm{cov}}
\newcommand{\eqdef}{\stackrel{\text{def}}{=}}
\newcommand{\y}{\textbf{y}}
\newcommand{\X}{\textbf{X}}

\def\<#1,#2>{\left\langle #1,#2\right\rangle}
\icmltitlerunning{A Self-supervised Approach to Hierarchical Forecasting with Applications to Groupwise Synthetic Controls}

\begin{document}

\twocolumn[
\icmltitle{A Self-supervised Approach to Hierarchical Forecasting\\ with Applications to Groupwise Synthetic Controls}



\icmlsetsymbol{equal}{*}

\begin{icmlauthorlist}
\icmlauthor{Konstantin Mishchenko}{kaust,amz}
\icmlauthor{Mallory Montgomery}{amz}
\icmlauthor{Federico Vaggi}{amz}
\end{icmlauthorlist}

\icmlaffiliation{kaust}{King Abdullah University of Science and Technology}
\icmlaffiliation{amz}{Amazon}

\icmlcorrespondingauthor{Konstantin Mishchenko}{konstmish.github.io}

\icmlkeywords{Machine Learning, ICML, Time Series}

\vskip 0.3in
]
\printAffiliationsAndNotice{}

\begin{abstract}
    When forecasting time series with a hierarchical structure, the existing state of the art is to forecast each time series independently, and, in a post-treatment step, to reconcile the time series in a way that respects the hierarchy~\cite{hyndman2011optimal, hyndman2018trace}.  We propose a new loss function that can be incorporated into any maximum likelihood objective with hierarchical data, resulting in reconciled estimates with confidence intervals that correctly account for additional uncertainty due to imperfect reconciliation.  We evaluate our method using a non-linear model and synthetic data on a counterfactual forecasting problem, where we have access to the ground truth and contemporaneous covariates, and show that we largely improve over the existing state-of-the-art method.  
\end{abstract}

\section{Introduction}

\begin{figure}
        \includegraphics[scale=.6, trim={0 1.41cm 0 1.39cm},clip]{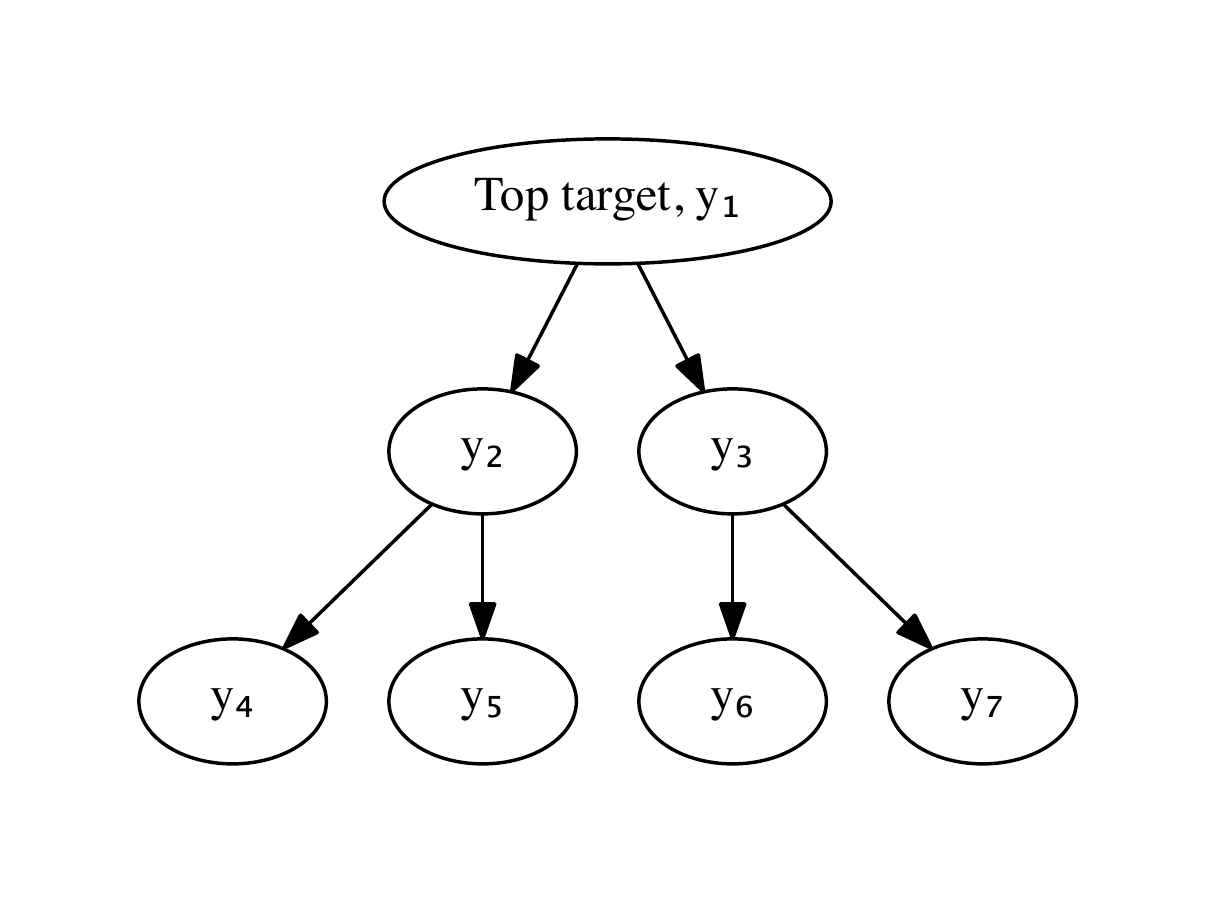}
        \caption{An example of a two-level hierarchy, in which $y_1=y_2+y_3$, $y_2=y_4+y_5$, $y_3 = y_6 + y_7$.}
        \label{fig:example_hierarchy}
    \end{figure}

Forecasting problems frequently have natural hierarchies. For example, looking at employment levels in the United States, state-level figures must sum to national-level, and industry-level must sum to overall employment. In Figure \ref{fig:example_hierarchy}, $y_4$ through $y_7$ could be counties that sum to the states $y_2$ and $y_3$, which sum to the national employment $y_1$.

Forecasting each time series independently leads to several undesirable outcomes. It throws away information on the nested structure of the problem, which can result in less accurate forecasts and unreliably estimated confidence intervals. Researchers estimating many time series simultaneously test multiple hypotheses about the nested forecasts, requiring them to adjust confidence intervals, diluting the statistical power of models. Further, discrepancies make decision-making more difficult for policymakers, who may not know how to translate a case where the child-level estimates add up to more than the parent-level forecast into good policy.

When forecasting time series with a known hierarchical structure, the current standard for achieving coherence in a given hierarchy is by  \cite{hyndman2011optimal, hyndman2018trace}, who propose an approach to combine independent forecasts for each time series to obtain a weighted sum that minimizes the reconciliation error. Given a set of forecasts $\hat{y}_i$ and a hierarchy $S$, Hyndman's ``optimal reconciliation'' (often referred to as HTS for ``hierarchical time series'') estimates a matrix $P$ such that $\tilde{y}_i = SP\hat{y}_i$, where $\tilde{y}_i$ are the reconciled forecasts. They use the forecasts and estimated variances of each time series to create simple prediction intervals that ignore the hierarchical structure. \cite{hyndman2018trace} derive an estimator for $P$ based on the trace of the covariance matrix of the forecast errors. 

We propose a different approach, leveraging techniques from the machine learning literature in self-supervised learning (SSL), which utilizes the structure of the problem to improve predictions when limited labeled data is available. In our case, the forecaster has access to labeled data from the training period alongside unlabeled data from the forecast period by relying on structural properties of the data. This approach has been used successfully in fields such as computer vision. In ~\cite{rasmus2015semi}, the authors propose that the classifications of perturbed (e.g., rotated) versions of the same image should be \textit{similar} to one another, even if the original is not labeled and thus we do not know whether they are correct.

Compared to HTS, our method results in more accurate forecasts (as measured by out-of-sample mean squared errors in simulations) that are closely reconciled. Because all forecasts are estimated simultaneously, our estimated confidence intervals account for uncertainty arising from reconciliation error, and avoid ex-post adjustments for multiple hypothesis testing.

When the child time series are heterogeneous, and the researcher has access to high-quality covariates that predict the parent and child time series well, disaggregating a single parent time series into multiple child time series and forecasting the entire hierarchy at once can lead to significant improvements in test accuracy, as shown in section \ref{section:Synthetic Data}. Randomly or arbitrarily splitting the parent into smaller time series will not improve test errors, and may lead to randomly divergent forecasts. For example, splitting a sample of companies alphabetically will be less useful than splitting them by size, location, or industry, for which we can find suitable theory-driven covariates.

This method is applicable to pure forecasting (with access only to past observations of the target time series and covariates) as well as to counterfactual forecasting (with access to contemporaneous covariate time series). In fact, this method can be used in any case with a maximum likelihood objective. Here we focus on counterfactual forecasting using a synthetic control method, initially developed by \cite{abadie2010synthetic}. They construct a counterfactual forecast for a US state with a tobacco tax increase by constructing an artificial control group using unaffected states as covariates in a pre-treatment window and creating a weighted sum.  In the original paper, the synthetic control was limited to a nonnegative weighted sum of other covariates (also called donors) and sparsity between the donors was enforced by choosing coefficients that were on the simplex (i.e., they sum to one).  These restrictions were lifted in subsequent work \cite{doudchenko2016balancing}, allowing for more flexible relationships between target and donors.  Once an artificial control group has been built, the effect of the intervention is defined as the difference between the control group and the treated group after the intervention. In the absence of an intervention, the synthetic control forecast should closely match the observed values.

\cite{BrodersenCausalImpact} proposed an extension to the synthetic control framework by using Bayesian structural time series models \cite{scott2013predicting} to form the synthetic control group.  By leveraging the flexibility of structural time series models, they can incorporate components into their counterfactual such as seasonality, covariates, and trend terms. There are many varieties of synthetic control methods, including generalized synthetic controls\cite{xuGsynth}, any of which can be estimated and reconciled using our proposed loss function.

\section{Theory} \label{Theory}
\textbf{Notation:} We denote target time series as $\y_i = (y_i^1,\dotsc, y_i^T)^\top\in \RR^T$ for $i=1,\dotsc, n$. For all $i$, we write the forecast for $\y_i$ as $\hat \y_i = (\hat y_i^1, \dotsc, \hat y_i^T)^\top$.

We now describe the theoretical basis for our estimations, beginning with the objective function we are interested in minimizing for the simple case of a hierarchy of $n$ time series, with $n-1$ children that sum to a single parent. (Additional levels of the hierarchy are estimated and reconciled in the same way. Note that we do not require the hierarchy to be structured as a tree -- we can optimize the model as long as the hierarchy can be described using a DAG.\footnote{For example, using overlapping product families that must simultaneously reconcile -- `popular categories for kids' and `popular categories for teens' may both contain apparel and school supplies. Our method forecasts and reconciles all categories and groupings simultaneously.}) The objective function is:
\begin{align}
    \min_{\hat \y_i} \overbrace{\sum_{i=1}^n \sum_{t=1}^{t_0}l\left( \hat y_i^t, y_i^t \right)}^{\text{forecasting loss}} + \overbrace{\lambda \sum_{t=t_0+1}^T \Bigl\|\hat y_1^t - \sum_{i=2}^n \hat y_i^t \Bigr\|^2}^{\text{reconciliation loss}}, \label{eq:mle_objective}
\end{align}
where $l(\cdot, \cdot)$ is a loss function, $\{y_i^t\}_{t=1}^{t_0}$ are the observations in the training period, $\{\hat y_i^t\}_{t=1}^T$ are forecasts in the test period of the $i$-th target
and $\lambda$ is the reconciliation penalty size. Without loss of generality, we will assume that $l(\hat y, y)=(\hat y - y)^2$ and that $\hat y_i^t(\theta)$ is a parametric forecast.

\begin{theorem}
    Assume that problem~\eqref{eq:mle_objective} has solutions for all $\lambda\ge 0$. Fix $\lambda> 0$ and let $\theta_\lambda^*$ be an approximate solution of~\eqref{eq:mle_objective} with functional suboptimality not greater than $\varepsilon$. Then, we have for the corresponding outputs $\hat \y_1,\dotsc, \hat\y_n$
    \begin{align}
        \sum_{t=t_0+1}^T \Bigl\|\hat y_1^t(\theta) - \sum_{i=2}^n \hat y_i^t(\theta) \Bigr\|^2
        = O\Bigl(\frac{1}{\lambda}\Bigr) + \frac{\varepsilon}{\lambda} \label{eq:generalization_bounds}
    \end{align}
\end{theorem}
\textit{\textbf{Proof}: See Appendix Section \ref{Appendix:theoremProof}.}

If $\lambda=0$, forecasts for each target $y_i$ become independent and are simply equal to the minimizers of the forecasting loss function. For any $\lambda>0$, each forecast will be adjusted to ensure 
$\|\hat y_1^t - \sum_{i=2}^n \hat y_i^t\|^2 = O(\frac{1}{\lambda})$
. Note that the reconciliation penalty only applies for $t>t_0$, because, for $t \leq t_0$, minimizing the forecasting loss already implicitly minimizes the reconciliation loss, as the ground truth data $y_i^t$ satisfies the hierarchy by construction.

This approach is applicable to any method of predicting $y_i^t$. If $y_i^t$ is a differentiable parametric function $f(\theta_i)$, we can optimize objective~\eqref{eq:mle_objective} using gradient based methods. In Appendix~ \ref{Appendix:Optimization algorithm} we provide an efficient algorithm to solve~\eqref{eq:mle_objective}.

We can view the reconciliation penalty as penalizing the forecaster if the time series start deviating from their natural hierarchy, and we can also justify it rigorously.
Ideally, we would like to find parameters that minimize test error $(\hat y_i^t - y_i^t)^2$ for $i=1,\dotsc,n$, but we do not have access to $y_i^t$ for $t > t_0$. However, the reconciliation penalty serves as a lower bound for the sum of test errors. Namely, since 
$y_1^t = \sum_{i=2}^n y_i^t$
, the Cauchy-Schwarz inequality implies
\begin{align*}
    \Bigl(\hat y_1^t - \sum_{i=2}^n \hat y_i^t\Bigr)^2
    &= \Bigl(\hat y_1^t - y_1^t - \sum_{i=2}^n (\hat y_i^t - y_i^t) \Bigr)^2\\
    &\le \sumin 1^2 \sumin(\hat y_i^t - y_i^t)^2 \\
    &= n \sumin(\hat y_i^t - y_i^t)^2.
\end{align*}

Intuitively, having perfect reconciliation is a necessary but not sufficient condition for a perfect forecast, therefore, if a hierarchical forecast does not perfectly reconcile, we can use that information to quantify the minimum amount by which the forecast is off.

\subsection{Bayesian modeling} \label{Bayesian modeling}
We now show that we can cast reconciliation in a Bayesian formulation, and use it to obtain confidence intervals that automatically incorporate uncertainty from the reconciliation step. We focus our attention on the case of counterfactual forecasting, where we have highly predictive covariates $\X_i$ available, and thus use a simple linear model for the likelihood. For a review of Bayesian statistics, including in time series, see \cite{GewekeTextbook}.

We forecast each time series $\y_i$ using a potentially disjoint set of covariates $\X_i$, and estimate a joint likelihood for all observations:
\begin{align}
    (y_1^t, \dotsc, y_n^t) &\mid X_1^t,\dotsc, X_n^t , \theta_1,\dotsc, \theta_n, \sigma \sim \cN(\mu^t, \Sigma)  \notag\\
    & \text{with }
    \mu^t \eqdef (X_1^t\theta_1, \dotsc, X_n^t\theta_n)^\top \label{eq:bayesian_train_likelihood}\\
    & \text{and LKJ prior on }
    \Sigma \sim \text{LKJcorr}(\omega_\Sigma, \eta), \notag
\end{align}
where LKJ prior is standard choice of matrix distribution for covariance matrix from~\cite {lkj_priors}, $\eta$ is a parameter that controls how uniform the non-diagonal terms are, and $\omega_\Sigma$ is a standard half-Cauchy prior on the diagonal elements of the covariance.

For the reconciliation term, note that, by definition, the difference between jointly Gaussian random variables is itself Gaussian, and therefore, for a simple hierarchy with a single parent $y_1$ we have:

\begin{align}
    \Bigl \{ \{y_1^t\}_{t=t_0+1}^T -  \sum_{i=2}^n \{y_i^t\}_{t=t_0+1}^T \Bigr \}\sim \cN(0, \sigma_{rec}
    ), 
\end{align}
with prior:
\begin{align*}
    \sigma_{rec} &\sim \text{half-Cauchy}\Bigl(\frac{1}{\lambda_{rec}}\Bigr).
\end{align*}\label{eq:bayesian_reconciliation}
In this formulation, the prior on the reconciliation variance $\lambda_{rec}$ plays  the same role as the $\lambda$ hyperparameter in \eqref{eq:mle_objective}.  If we choose a large value for $\lambda_{rec}$, we are expressing our prior belief that forecasts should reconcile closely.

An alternative approach to estimating $\sigma_{rec}$ is to observe that every $y_i$ is a random variable, and therefore, the linear combination $y_1^t - \sum_{i=2}^n y_i^t$ is random as well. For $t>t_0$, its variance by definition is equal to
\begin{align}
    &\Var\Bigl(y_1^t - \sum_{i=2}^n y_i^t\Bigr) \notag\\
    &= \sumin \Var(y_i^t) - 2\sum_{i=2}^n \cov(y_1^t, y_i^t) + \sum_{i,j=2}^n \cov (y_i^t, y_j^t) \nonumber\\
    &= \sumin \Sigma_{i,i}  - 2\sum_{i=2}^n \Sigma_{1,i} + \sum_{\substack{i,j=2\\ i\neq j}}^n \Sigma_{i, j}. \label{eq:reconciliation_variance}
\end{align} 

Therefore, when we have the full covariance matrix of the forecast errors $\Sigma$, we can derive $\sigma_{rec}$ analytically.  One property of this approach is that by specifying an additional likelihood for each reconciliation step, we avoid unrealistically narrow confidence intervals: if the model has learned a very accurate forecast for every target variable, the variance in the likelihood of the reconciliation term will necessarily be small as well, as it is a linear function of the forecast variances.  This means that the model cannot overfit to the training set and predict unreasonably tight confidence intervals, as those predictions will give a large penalty for non-reconciliation.  By jointly estimating the reconciliation and forecast likelihoods, we force the model to come to terms with what it does not know, which translates into decreased confidence in the forecast.

\section{Applications}\label{section:Applications}

\subsection{Synthetic data}\label{section:Synthetic Data}
Synthetic control models are typically used to estimate the causal impact of a treatment, but there need not be an intervention. Here, we use synthetic data with no intervention, but that includes predictive contemporaneous covariates.
\begin{table*}[t]
\center
\scriptsize
\begin{tabular}{lll|c|c|c|c}
\toprule
      &     &       &   \emph{no\_hierarchy}            &    \emph{no\_hierarchy + hts}      & \emph{full\_hierarchy\_lam\_1} & \emph{full\_hierarchy\_lam\_10} \\
\midrule
\textbf{test} & \textbf{mse} & y\_1 &          5.78 $\pm$ 12.7 &          5.78 $\pm$ 12.8 &          1.62 $\pm$ 2.56 &           \textbf{1.61 $\pm$ 2.72} \\
      &     & y\_2 &          3.02 $\pm$ 9.36 &          3.02 $\pm$ 9.44 &         0.894 $\pm$ 1.59 &           \textbf{0.89 $\pm$ 1.39} \\
      &     & y\_3 &          2.79 $\pm$ 5.02 &          2.78 $\pm$ 5.01 &          1.03 $\pm$ 2.13 &          \textbf{0.956 $\pm$ 1.85} \\
      &     & y\_4 &           1.29 $\pm$ 4.1 &            1.3 $\pm$ 4.0 &        0.496 $\pm$ 0.874 &         \textbf{0.462 $\pm$ 0.692} \\
      &     & y\_5 &          1.06 $\pm$ 2.1 &           1.07 $\pm$ 2.09 &         \textbf{0.42 $\pm$ 0.673} &         0.425 $\pm$ 0.703 \\
      &     & y\_6 &           1.1 $\pm$ 1.92 &           1.1 $\pm$ 1.94 &         0.501 $\pm$ 1.17 &         \textbf{0.452 $\pm$ 0.826} \\
      &     & y\_7 &          1.09 $\pm$ 1.8 &            1.1 $\pm$ 1.78 &        0.445 $\pm$ 0.638 &         \textbf{0.422 $\pm$ 0.609} \\
\hline
      & \textbf{rec} & total &   0.0392 $\pm$ 0.0668  &     \textbf{4.62e-32 $\pm$ 1e-31}   &  3.28e-05 $\pm$ 1.66e-05 &    1.5e-05 $\pm$ 6.81e-06 \\
\hline
\textbf{train} & \textbf{mse} & y\_1 &    \textbf{0.00575 $\pm$ 0.00518} &  \textit{*}   &      0.0212 $\pm$ 0.0205 &       0.0265 $\pm$ 0.0253 \\
      &     & y\_2 &    \textbf{0.00414 $\pm$ 0.00408} &     &      0.0108 $\pm$ 0.0104 &       0.0134 $\pm$ 0.0122 \\
      &     & y\_3 &   \textbf{0.00414 $\pm$ 0.00397} &     &      0.0109 $\pm$ 0.0105 &       0.0135 $\pm$ 0.0128 \\
      &     & y\_4 &    \textbf{0.00311 $\pm$ 0.00318} &     &    0.00562 $\pm$ 0.00543 &     0.00684 $\pm$ 0.00632 \\
      &     & y\_5 &    \textbf{0.00311 $\pm$ 0.00305} &     &    0.00563 $\pm$ 0.00543 &      0.0068 $\pm$ 0.00627 \\
      &     & y\_6 &    \textbf{0.00311 $\pm$ 0.00307} &     &     0.00562 $\pm$ 0.0054 &      0.00685 $\pm$ 0.0063 \\
      &     & y\_7 &     \textbf{0.0031 $\pm$ 0.00304} &      &    0.00567 $\pm$ 0.00551 &     0.00696 $\pm$ 0.00657 \\
\hline
      & \textbf{rec} & total &  \textbf{0.000195 $\pm$ 0.000118} &   &  0.000311 $\pm$ 0.000141 &   0.000817 $\pm$ 0.000346 \\
\bottomrule
\end{tabular}
\caption{The first column, \emph{no\_hierarchy} shows results from an independently estimated, unreconciled model, while the next column, \emph{no\_hierarchy + hts} uses the same forecasts, reconciled using Hyndman's HTS. Both \emph{full\_hierarchy\_lam\_1} and \emph{full\_hierarchy\_lam\_10} use our method with $\lambda = 1$ and $\lambda = 10$ respectively. \textit{* We omit train MSE numbers in the second column because they are identical to the unreconciled figures.}}
\end{table*}\label{table:reconciliation-synthetic-data}

To assess the extent to which our reconciliation approach works for non-linear models, we consider an artificial dataset with ground truth available, with $m$ covariates and $n$ target variables, as in the hierarchy from Figure \ref{fig:example_hierarchy}. In this setting, we show that reconciliation serves as an effective regularizer, improving forecasts even compared to HTS.

Each covariate time series $\X_i$ is drawn from a Gaussian process prior using the Celerite library~\cite{foreman2017fast} and each leaf target variable $y_i^t$ is defined as a linear transformation of both $(X_1^t, \dotsc, X_m^t)$ and $(X_1^t\cdot t, \dotsc, X_m^\cdot t)$ plus random noise.  For non-leaf target variables, we sum up the children using the hierarchy shown in Figure~\ref{fig:example_hierarchy}. Formally, we take a kernel $k(\cdot, \cdot)$ from Celerite and sample
\begin{align*}
    X_j^t &\sim GP(0, k(X_j^t, X_j^{t^{'}})),\quad j=1,\dotsc, m \\
    y_i^t &\eqdef (X_1^t\theta_1 + (X_1^t \cdot t) \phi_1 , \dotsc, X_m^t\theta_m + (X_m^t \cdot t) \phi_m + \epsilon)^\top.
\end{align*}

We then use a multi-layer perceptron (one hidden layer with 100 units and ReLU activations) to forecast $Y$ using $X$ as an input.  The neural network gets the first 1000 time steps as a training set, and we then report the MSE (averaged over 1000 independent experiments) on train, test, and reconciliation in table ~\ref{table:reconciliation-synthetic-data}.

We trained the neural networks without a reconciliation penalty (\emph{no\_hierarchy}), and with two different values of $\lambda$, 1 and 10 (\emph{full\_hierarchy\_lam\_1} and \emph{full\_ hierarchy\_lam\_10}).  We also used HTS (conjugate gradient method) to reconcile the forecasts from the \emph{no\_hierarchy}) baseline and report those values in the \emph{no\_hierarchy + hts} column.

Reconciliation serves as a regularizer: by adding the reconciliation penalty (\emph{full\_hierarchy\_lam\_1} and \emph{full\_hierarchy\_lam\_10} columns), we decrease train set performance (4x increase in train MSE), but improve test set performance, reducing MSE by up to 3.5x compared to the \emph{no\_hierarchy} baseline in the parent node $\y_1$.  Although the reconciliation penalty decreases test set error for all nodes, the reduction is greater on nodes higher in the hierarchy ($\y_1$, $\y_2$, and $\y_3$).  The decrease in test error is far greater than the decrease in reconciliation error; we believe this is because, by jointly training with a reconciliation loss, we give the model an inductive bias toward simpler solutions with lower test error.  In contrast, HTS adjusts the forecasts as little as possible after training is complete in order to achieve reconciliation ex-post, so it can at most reduce the forecast error by the amount of reconciliation error, which is much smaller. This is why HTS reduces the reconciliation error to approximately zero, but with little improvement in test forecast accuracy.

\section{Conclusion}
This paper focuses on counterfactual forecasting, however, the reconciliation loss can be applied to any maximum likelihood objective with hierarchical data, including multi-target regression and pure forecasting.  When estimating hierarchical times series using e.g. ARIMA, we can add a reconciliation term to the maximum likelihood objective and automatically obtain forecasts that are nearly reconciled and have more sensible uncertainty estimates.

Although we only prove lower bounds on the error term, we see that applying our reconciliation loss to synthetic data, where we have access to real ground truth, forecasts with the reconciliation loss term have higher accuracy compared to a strong HTS baseline.  In future work, we are planning a more systematic investigation into how reconciliation affects the error as well as the inductive bias of the model.

\bibliography{REMI}

\begin{thebibliography}{12}
\providecommand{\natexlab}[1]{#1}
\providecommand{\url}[1]{\texttt{#1}}
\expandafter\ifx\csname urlstyle\endcsname\relax
  \providecommand{\doi}[1]{doi: #1}\else
  \providecommand{\doi}{doi: \begingroup \urlstyle{rm}\Url}\fi

\bibitem[Abadie et~al.(2010)Abadie, Diamond, and
  Hainmueller]{abadie2010synthetic}
Abadie, A., Diamond, A., and Hainmueller, J.
\newblock Synthetic control methods for comparative case studies: Estimating
  the effect of california’s tobacco control program.
\newblock \emph{Journal of the American statistical Association}, 105\penalty0
  (490):\penalty0 493--505, 2010.

\bibitem[Brodersen et~al.(2015)Brodersen, Gallusser, Koehler, Remy, and
  Scott]{BrodersenCausalImpact}
Brodersen, K.~H., Gallusser, F., Koehler, J., Remy, N., and Scott, S.~L.
\newblock Inferring causal impact using bayesian structural time-series models.
\newblock \emph{Annals of Applied Statistics}, 9:\penalty0 247--274, 2015.

\bibitem[Doudchenko \& Imbens(2016)Doudchenko and
  Imbens]{doudchenko2016balancing}
Doudchenko, N. and Imbens, G.~W.
\newblock Balancing, regression, difference-in-differences and synthetic
  control methods: A synthesis.
\newblock Technical report, National Bureau of Economic Research, 2016.

\bibitem[Foreman-Mackey et~al.(2017)Foreman-Mackey, Agol, Ambikasaran, and
  Angus]{foreman2017fast}
Foreman-Mackey, D., Agol, E., Ambikasaran, S., and Angus, R.
\newblock Fast and scalable gaussian process modeling with applications to
  astronomical time series.
\newblock \emph{The Astronomical Journal}, 154\penalty0 (6):\penalty0 220,
  2017.

\bibitem[Geweke(2005)]{GewekeTextbook}
Geweke, J.
\newblock \emph{Contemporary Bayesian Econometrics and Statistics}.
\newblock Wiley, 2005.

\bibitem[Hyndman et~al.(2011)Hyndman, Ahmed, Athanasopoulos, and
  Shang]{hyndman2011optimal}
Hyndman, R.~J., Ahmed, R.~A., Athanasopoulos, G., and Shang, H.~L.
\newblock Optimal combination forecasts for hierarchical time series.
\newblock \emph{Computational Statistics \& Data Analysis}, 55\penalty0
  (9):\penalty0 2579--2589, 2011.

\bibitem[Lewandowski et~al.(2009)Lewandowski, Kurowicka, and Joe]{lkj_priors}
Lewandowski, D., Kurowicka, D., and Joe, H.
\newblock Generating random correlation matrices based on vines and extended
  onion method.
\newblock \emph{Journal of multivariate analysis}, 100\penalty0 (9):\penalty0
  1989--2001, 2009.

\bibitem[Mishchenko \& Richt{\'a}rik(2018)Mishchenko and
  Richt{\'a}rik]{mishchenko2018stochastic}
Mishchenko, K. and Richt{\'a}rik, P.
\newblock A stochastic penalty model for convex and nonconvex optimization with
  big constraints.
\newblock \emph{arXiv preprint arXiv:1810.13387}, 2018.

\bibitem[Rasmus et~al.(2015)Rasmus, Berglund, Honkala, Valpola, and
  Raiko]{rasmus2015semi}
Rasmus, A., Berglund, M., Honkala, M., Valpola, H., and Raiko, T.
\newblock Semi-supervised learning with ladder networks.
\newblock In \emph{Advances in Neural Information Processing Systems}, pp.\
  3546--3554, 2015.

\bibitem[Scott \& Varian(2013)Scott and Varian]{scott2013predicting}
Scott, S.~L. and Varian, H.~R.
\newblock Predicting the present with bayesian structural time series.
\newblock \emph{Available at SSRN 2304426}, 2013.

\bibitem[Wickramasuriya et~al.(2018)Wickramasuriya, Athanasopoulos, and
  Hyndman]{hyndman2018trace}
Wickramasuriya, S.~L., Athanasopoulos, G., and Hyndman, R.~J.
\newblock Optimal forecast reconciliation for hierarchical and grouped time
  series through trace minimization.
\newblock \emph{Journal of the American Statistical Association}, 0\penalty0
  (0):\penalty0 1--16, 2018.
\newblock \doi{10.1080/01621459.2018.1448825}.
\newblock URL \url{https://doi.org/10.1080/01621459.2018.1448825}.

\bibitem[Xu(2017)]{xuGsynth}
Xu, Y.
\newblock Generalized synthetic control method: Causal inference with
  interactive fixed effects models.
\newblock \emph{Political Analysis}, 25:\penalty0 57--76, 2017.
\newblock \doi{10.1017/pan.2016.2}.

\end{thebibliography}
\bibliographystyle{icml2019}
\clearpage
\onecolumn
\appendix
\section{Proof of theorem 1}\label{Appendix:theoremProof}
The analysis below is motivated by that of constrained optimization with penalties~\cite{mishchenko2018stochastic}.
\begin{proof}
    The objective~\eqref{eq:mle_objective} can be also seen as a penalty reformulation of the harder problem:
    \begin{equation}
    \begin{aligned}
        &\min_{\hat \y_i} && \sum_{i=1}^n \sum_{t=1}^{t_0}l\left( \hat y_i^t, y_i^t \right)\\
        &\text{subject to} && \hat y_1^t = \sum_{i=2}^n \hat y_i^t,\enskip t=t_0 + 1, \dotsc, T.
    \end{aligned}\label{eq:constr_problem}
    \end{equation}

    Let $\theta^*$ be a solution of~\eqref{eq:constr_problem}, and denote by $P_\lambda(\cdot)$ the objective function in~\eqref{eq:mle_objective}. We plug $\theta^*$ into~\eqref{eq:mle_objective} and by the definition of $\theta_\lambda^*$ we have
    \begin{align*}
        P_\lambda(\theta_\lambda^*) \le \min_\theta P_\lambda(\theta) + \varepsilon \le P_\lambda(\theta^*) + \varepsilon.
    \end{align*}
    Moreover, for $\lambda=0$ we get by definition of $\theta_0^*$ that $P_0(\theta_0^*) \le P_0(\theta_\lambda^*)$. Noting that 
    \begin{align*}
        P_\lambda(\theta) 
        = P_0(\theta) + \lambda\sum_{t=t_0+1}^T \left\|\hat y_1^t(\theta) - \sum_{i=2}^n \hat y_i^t(\theta) \right\|^2,
    \end{align*}
    we deduce $P_0(\theta_\lambda^*) \le P_\lambda(\theta_\lambda^*)$. Since $\theta^*$ is a feasible solution of~\eqref{eq:constr_problem}, we additionally get $P_\lambda(\theta^*) = P_0(\theta^*)$. Therefore,
    \begin{align*}
        \lambda\sum_{t=t_0+1}^T \left\|\hat y_1^t(\theta^*) - \sum_{i=2}^n \hat y_i^t(\theta^*) \right\|^2
        \le P_\lambda(\theta^*)  - P_\lambda(\theta_\lambda^*) + \varepsilon
        \le P_0(\theta^*) - P_0(\theta_0^*) + \varepsilon.
    \end{align*}
    Dividing both sides by $\lambda$ finishes the proof.
\end{proof}
\section{Optimization algorithm} \label{Appendix:Optimization algorithm}
\begin{algorithm}[h]
  \caption{Randomized block coordinate descent for~\eqref{eq:mle_objective}}
  \label{alg:rcd}
\begin{algorithmic}[1]
\STATE \textbf{Input: }{initial parameter vectors $\theta_1^0, \dotsc, \theta_n^0$, stepsizes $\gamma_1, \dotsc, \gamma_n$, penalty size $\lambda>0$, forecast model $\hat y_i^t = f(X^t; \theta_i)$, number of iterations $K$}
    \FOR{$k=0,\dotsc, K$}
        \STATE Sample $i$ uniformly from $\{1,\dotsc,n\}$
        \STATE Compute partial gradients $g_i^k = \sum_{t=1}^{t_0}\frac{\partial}{\partial \theta_i} l(f(\X_i^t; \theta_i), y_i^t) + 2\lambda \sum_{t=t_0+1}^T \frac{\partial}{\partial \theta_i}f(\X_i^t; \theta_i) \left\| \hat y_1^t - \sum_{j=2}^n \hat y_j^t  \right\|$
        \STATE Update parameters $\theta_i^{k+1} = \theta_i^k - \gamma_i g_i^k$ and $\theta_j^{k+1} = \theta_j^k$ for $j\neq i$
        \STATE Update forecasts $\hat y_i^t = f(\X_i^t; \theta_i)$ for $t=1,\dotsc, T$
    \ENDFOR
\end{algorithmic}
\end{algorithm}
Problems~\eqref{eq:mle_objective} and \eqref{eq:constr_problem} can be solved efficiently using gradient methods. If the model used to produce $\hat y_i^t$ is simple, we can use projected gradient methods for~\eqref{eq:constr_problem}. Its relaxed version~\eqref{eq:mle_objective}, however, is easier since we can compute the derivatives of the penalty term. Furthermore, it allows for using more efficient coordinate descent optimizers such as Algorithm~\ref{alg:rcd}.

The main feature of randomized coordinate descent applied to~\eqref{eq:mle_objective} is that it automatically randomizes data sampling as well. Indeed, if we are adjusting the parameters used to produce $\hat y_i^t$, $t=1,\dotsc, T$, all terms that use $y_j^t$ for any $t$ and $j\neq i$ are not updated and, thus, we do not have to look up the related data. This means that the update is $n$ times more efficient in how many coordinates it updates and, additionally, up to $n$ times more efficient in how much data it processes. This makes overall training much faster.

However, we would like to note that with Bayesian model~\eqref{eq:bayesian_train_likelihood} we do not need to choose $\lambda$ if the adaptive rule~\eqref{eq:reconciliation_variance} is used. Since it requires estimating the covariance matrix, the rule from~\eqref{eq:reconciliation_variance} is much harder numerically, but would produce better estimates.

\end{document}